\pdfoutput=1

\documentclass[11pt]{article}

\usepackage{naacl2021}

\usepackage{times}
\usepackage{latexsym}
\usepackage{graphicx}

\usepackage[T1]{fontenc}

\usepackage[utf8]{inputenc}

\usepackage{microtype}
\usepackage{color, colortbl}
\usepackage{textgreek}
\usepackage{hyperref}
\usepackage{textcomp}
\usepackage{xcolor}
\usepackage{booktabs}
\usepackage{amssymb}
\usepackage{amsmath}

\title{Pose-Guided Sign Language Video GAN with Dynamic Lambda}

\author{Christopher Kissel, Christopher Kümmel, Dennis Ritter, Kristian Hildebrand\\
    Beuth University of Applied Sciences Berlin\\
}

\begin{document}
\maketitle
\begin{abstract}
    We propose a novel approach for the synthesis of sign language videos using GANs. We extend the previous work of Stoll et al.~\shortcite{Text2Sign} by using the human semantic parser of the Soft-Gated Warping-GAN from~\cite{WarpingGan} to produce photorealistic videos guided by region-level spatial layouts.
    Synthesizing target poses improves performance on independent and contrasting signers.
    Therefore, we have evaluated our system with the highly heterogeneous MS-ASL dataset~\cite{MSASL} with over 200 signers resulting in a SSIM of 0.893.
    Furthermore, we introduce a periodic weighting approach to the generator that reactivates the training and leads to quantitatively better results.
\end{abstract}

\begin{figure*}[h]
  \centering
  \includegraphics[width=\textwidth]{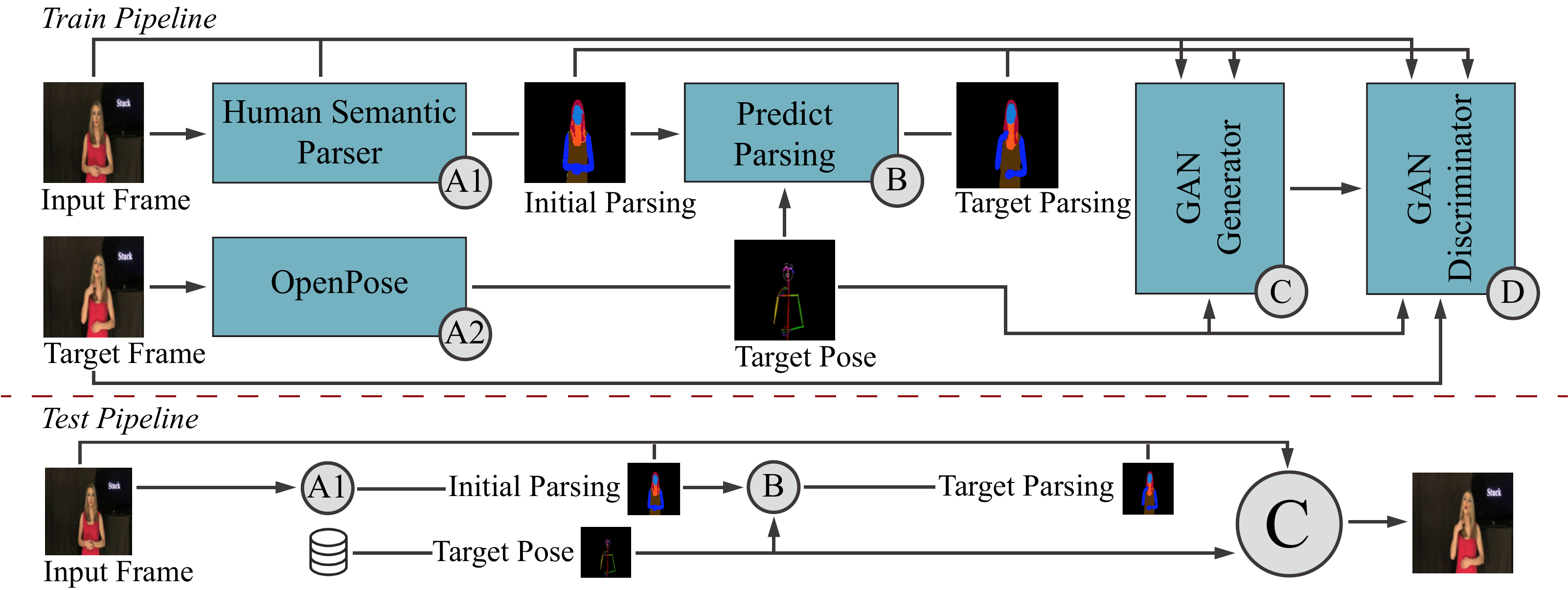}
  \caption{(Top:) During training, an input frame and a target frame are selected from a sign language video to serve as ground truth to train our models. The human semantic parser (A1) creates an initial parsing output of the input frame, while OpenPose (A2) extracts key joints from the target frame. These outputs are concatenated and the target parsing is predicted (B). Finally, input frame, initial and predicted parsing and key joints are used as input for generator (C). It generates an image that is as similar as possible to the target frame input. Whereas the discriminator (D) uses the initial frame, initial parsing, predicted parsing, target pose and the ground truth target frame to check the generated image of C for authenticity. (Bottom:) For testing, the input of C is the same as described above but the target pose is taken from prerecorded key joints. Finally, C creates an image of the person shown in the input frame in the desired pose.}
  \label{fig:pipeline}
\end{figure*}

\section{Introduction}
The WHO states that 430 million people in the world are deaf or suffer from significant hearing loss. Affected people often use one of the many different sign languages to communicate. In both directions of translation, the first approaches to barrier-free communication between signers and non-signers have emerged in recent years~\cite{bangham2000virtual, RWTH2014, camgoz1}. Our approach refers to the generation of personalized sign videos that can be synthesized in a pose-guided pipeline. Our network architecture is based on the findings of the Text2Sign~\cite{Text2Sign} paper, which presented the first non-graphical avatar approach to continuous gesture video generation. However, we only consider here the generation of glosses in videos. Our sign language GAN generates a video over the space of all signs from the MS-ASL dataset~\cite{MSASL} from a set of input-target pairs.
Our goal is to improve the approach by Stoll et al. and enhance the quality of the generated videos to support a variety of signers and different scene conditions.
Hence, we apply our approach to the challenging MS-ASL dataset. In this way, we hope to drive the development of real-world sign language applications with personalized generated signers.



When training on highly variable data, such as the MS-ASL dataset, we have observed that our approach first finds a stable equilibrium, but after some time the discriminator loss starts to increase, which means that it is too easily fooled (see Figure~\ref{fig:dynamic_vs_static}). It seems, however, that with a dataset as diverse as MS-ASL, not enough uniform criteria can be learned to judge the authenticity of an image when the generator has exceeded a certain threshold. If the discriminator is too weak, the L1 loss of the generator will prevail and the resulting images will be blurred. We have evaluated different strategies to stabilize the training and found that a dynamic lambda, a value that periodically weights the generator error against the discriminator, reactivates the training and leads to an improvement in image quality as shown in Figure~\ref{fig:static_vs_dynamic_result_variance}.

\section{Related Work}
Sign language related tasks can be divided into three subcategories
\emph{Sign Language Recognition (SLR)}~\citep{AlphabetRecognition, DeepASL, GestureRecognitionASL}, \emph{Sign Language Translation (SLT)}~\citep{camgoz1, camgoz2} and~\emph{Sign Language Production (SLP)}~\citep{Text2Sign}. Our approach relates to SLP tasks to produce glosses and signs of performed sign language.
Sign language datasets mostly consist of annotated videos. Recent approaches like "Word Level American Sign Language" (WLASL)~\citep{WLASL} or "Microsoft - American Sign Language" (MS-ASL)~\citep{MSASL} contain annotated high quality videos showing many signers performing various glosses. For our approach we use the MS-ASL dataset as it contains over 25000 videos of more than 200 signers with very different scene conditions, which comes closest to video data in the wild. Other datasets like SMILE~\citep{SMILE} used by~\cite{Text2Sign} or~\citep{how2sign} even contain multi-sensor data but fewer signers and unfortunately have not been published yet.
The use of GANs to tackle SLP problems is still rather new.
An important contribution from 2016 is the pix2pix~\citep{Pix2Pix} framework which uses a U-Net~\citep{UNet} generator with skip-connections.
It is a key component in our SLP approach as well. Synthesizing sign language videos share some challenges with tasks of Warping-GANs. They aim to transfer a person into previously unseen poses. There have been some remarkable developments in this area in the last years \citep{WarpingGan, Guha, PoseGuided, Liquid}.
In early 2020, Stoll et al. released a comprehensive approach to translate spoken language into sign language videos. The pipeline called \emph{Text2Sign}~\citep{Text2Sign} solves the task in two major stages. First the spoken glosses are translated into a sequence of motion graphs by an encoder-decoder architecture (Text2Pose). These poses are then passed to a sign language production GAN (Pose2Video), which is also based on a pix2pix architecture and an extension \citep{Pix2PixHD} of it. Our work picks up on the results of the \emph{Pose2Video} part and further develops the approach. Later in 2020, more GAN applications for SLP have been released \citep{multichannelSLP, SignSynth, caneverybody, WiGAN}. Which is an indication of the great interest in the issue and show improved results. Contrary to all this work, our approach concentrates on the synthesis of sign language in more real life conditions using the MS-ASL dataset.

\section{Method}
\label{sec:method}
\subsection{Approach \& Architecture}
Our \emph{SignLanguageGAN} pipeline combines the approaches from Soft-Gated Warping-GAN~\citep{WarpingGan} and Text2Sign~\cite{Text2Sign} and is displayed in Figure~\ref{fig:pipeline}. At the first stage the human semantic parser (A1), a U-Net~\citep{UNet} autoencoder trained on the CIHP dataset, a subset of the LIP dataset~\citep{LIP} creates an abstraction of the input frame. It has been shown that a semantic translation can be beneficial for spatial transformation of a person in an image \citep{WarpingGan, Guha}. The human semantic parser provides such a mapping from limb categories to colour values. This initial parsing is now transformed by a pix2pix network~\citep{Pix2Pix} (B) using the initial parsing and the OpenPose~\citep{OpenPose1} generated target pose (A2) representing the desired gloss. Finally, the result is the target parsing, a semantic color map of the person in the input frame who takes the pose represented by the OpenPose skeleton.
The second stage forms the actual adversarial model. Starting with a generator (C) that was expanded by a second convolutional encoder network. The latent spaces of both encoders are concatenated and a decoder learns to reproduce the target image from them. The encoder-decoder with skip connections is derived from U-Net architecture~\citep{UNet}, which was utilized in pix2pix. However, unlike in the original U-Net architecture, down-sampling is performed from two separate networks, while up-sampling is executed by a single one which merges the information of its counterparts. In its setup, the first encoder processes the generated target parsing from the first stage together with the OpenPose generated target pose. Encoder two, on the other hand, receives the input frame together with the respective initial parsing. Both encoders learn their individual mapping into a low dimensional representation. From the concatenation of these two latent spaces the decoder learns how to create a new image of the person seen in the initial frame performing the desired sign language gloss. This output is passed to the SignLanguageGAN discriminator (D) for the adversarial training. The discriminator determines whether the image originates from the generator or if a real sample from the dataset was used. In particular, it is important that the discriminator receives all information for assessment that was processed in the generation.

\subsection{Training with a Dynamic Lambda}
\begin{figure}[h]
  \includegraphics[width=0.48\textwidth]{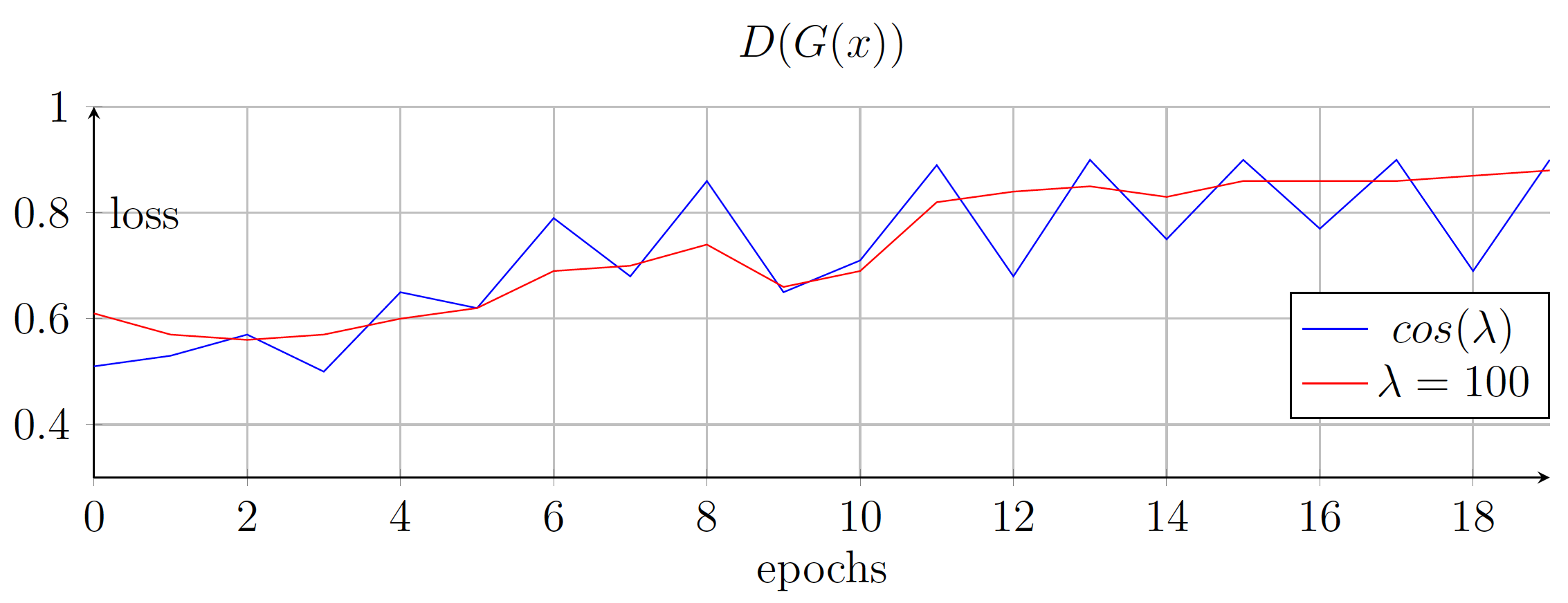}
  \caption{SignLanguageGAN discriminator losses for 20 epochs of training with dynamic lambda (blue) and static lambda (red).}
  \label{fig:dynamic_vs_static}
\end{figure}

While our architecture is a composite of several technologies, the training process differs in certain aspects from the frameworks used. We apply a least squares objective function in the discriminator as Mao et al. show that the L2 loss reduces vanishing gradients due to its sensitivity~\citep{LSGAN}.
This effect is particularly noticeable with highly heterogeneous data such as the MS-ASL~\citep{MSASL} collection.
Another challenge in working with such dynamic data became apparent as the training has advanced. The equilibrium between discriminator and generator becomes increasingly unstable, presumably because the discriminator has more and more difficulty learning uniform criteria to determine. This results in images with more artifacts and lower variance $\sigma$, as the generator contribution becomes stronger, for which we choose a L1 loss in accordance with pix2pix. $\sigma$ influences contrast and structure in SSIM. To break this trend, we apply a dynamic weighting between generator and discriminator to cyclically amplify the impact of the discriminator. For this we use the lambda parameter, which in pix2pix is a constant value. Several approaches have been evaluated to adjust lambda periodically or adaptively. However, strategies that interfered too much with the balance of the GAN were counterproductive. It have become evident that a steady fluctuation yielded the best results (see Figure~\ref{fig:dynamic_vs_static}).
Compared to a static weighting the dynamic lambda actively manipulates the loss leading to more updates for the generator.
Those updates encourage the generator to increase the output quality in order to fool the discriminator.
With the aim of having a high generator contribution at the beginning of the training, we use a cosine period with an amplitude span of 1.5 epochs $\approx \pi$ and a value range of 50 to 150. Our objective function consists of the conditional least squares discriminator objective $V(D,G)$ and the $L1$ loss with the dynamic term where $i$ and $j$ determine the amplitude range.

$$\lambda = i + \frac{cos([0, \pi]) + 1}{2} \cdot j$$
$$\min_{D}\min_{G}L_{SgnlGAN} = V(D,G)+L_{1}(G) \cdot \lambda$$

\begin{figure}[h]
  \includegraphics[width=0.48\textwidth]{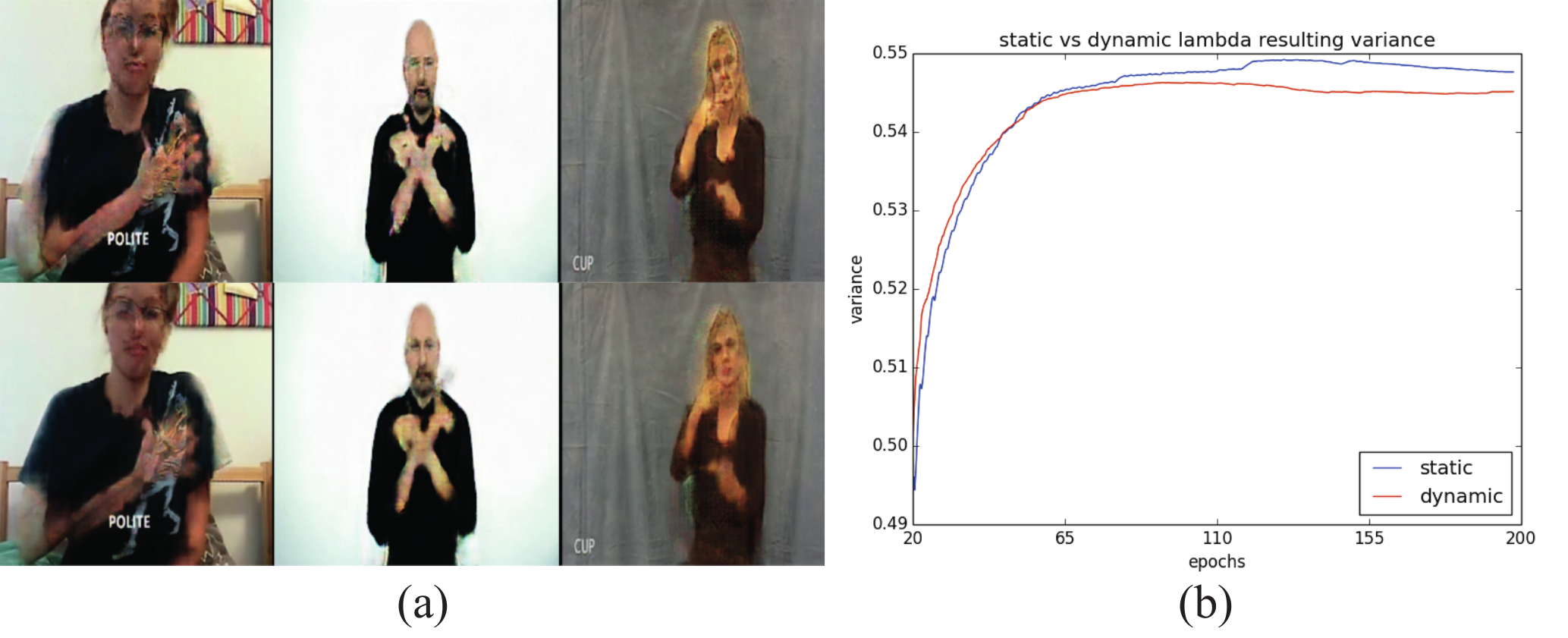}
  \caption{(a) (Top:) Static lambda during training. (Bottom:) Dynamic lambda during training. (b) Curve shows increased variance (in blue) when dynamic weighting is applied which leads to higher SSIM.}
  \label{fig:static_vs_dynamic_result_variance}
\end{figure}

By periodically manipulating the loss we observe an overall better performing generator with fewer artifacts and higher variance. This leads to a slight improvement in the SSIM and improved image results as shown in Figure~\ref{fig:static_vs_dynamic_result_variance}(a,b).




\section{Evaluation}

\subsection{Setup}


We evaluate our approach on the MS-ASL~\cite{MSASL} dataset, the most diverse sign language dataset in terms of number of signers and scene conditions. We compare our generated frames with ground truth images and determine the MSE, PSNR and SSIM to assess their quality. Note, unfortunately, we cannot compare our results directly to Text2Sign~\cite{Text2Sign}, since the SMILE dataset~\cite{SMILE} is not publicly available.

\begin{table}[h]
  \centering
  \begin{tabular}{ c | c | c | c }
    \toprule
    {Architecture}     & {PSNR}          & {MSE}            & {SSIM}         \\ \midrule
    pix2pix            & 20.546          & 783.376          & 0.786          \\
    \addlinespace[0.5em]
    Soft-Gated Warping & \textbf{22.033} & \textbf{622.683} & \textbf{0.813} \\
  \end{tabular}
  \caption{pix2pix and Soft-Gated Warping architectures tested on 50 videos of the MS-ASL dataset after three epochs of training.}
  \label{tab:architectures}
\end{table}

\subsection{Multi-Signer Generation Challenges}
In order to evaluate the Soft-Gated Warping extension in our architecture we have compared it with a network arrangement using pix2pix~\cite{Pix2Pix} instead. Table~\ref{tab:architectures} shows that the human semantic parser extension leads to improved results for all evaluated metrics using 50 randomly selected videos from the MS-ASL dataset. We also have compared these multi-signer MS-ASL results to results on homogenious \emph{Gebärdenlernen}\footnote{\url{https://gebaerdenlernen.de/index.php?article_id=72}} videos, which show a single-signer in the same scene performing german sign language glosses. In this way, we wanted to assess how the number of signers and the conditions of the scene affect SLP quality. The results show that single signer setups (with PSNR=25.634, MSE=194.174, SSIM=0.882) can easily outperform multi-signer datasets (with PSNR=22.033, MSE=622.683, SSIM=0.813). That underlines our assumption that our pose-guided approach would perform well in real-world scenarios.



\begin{figure}[h]
  \includegraphics[width=0.48\textwidth]{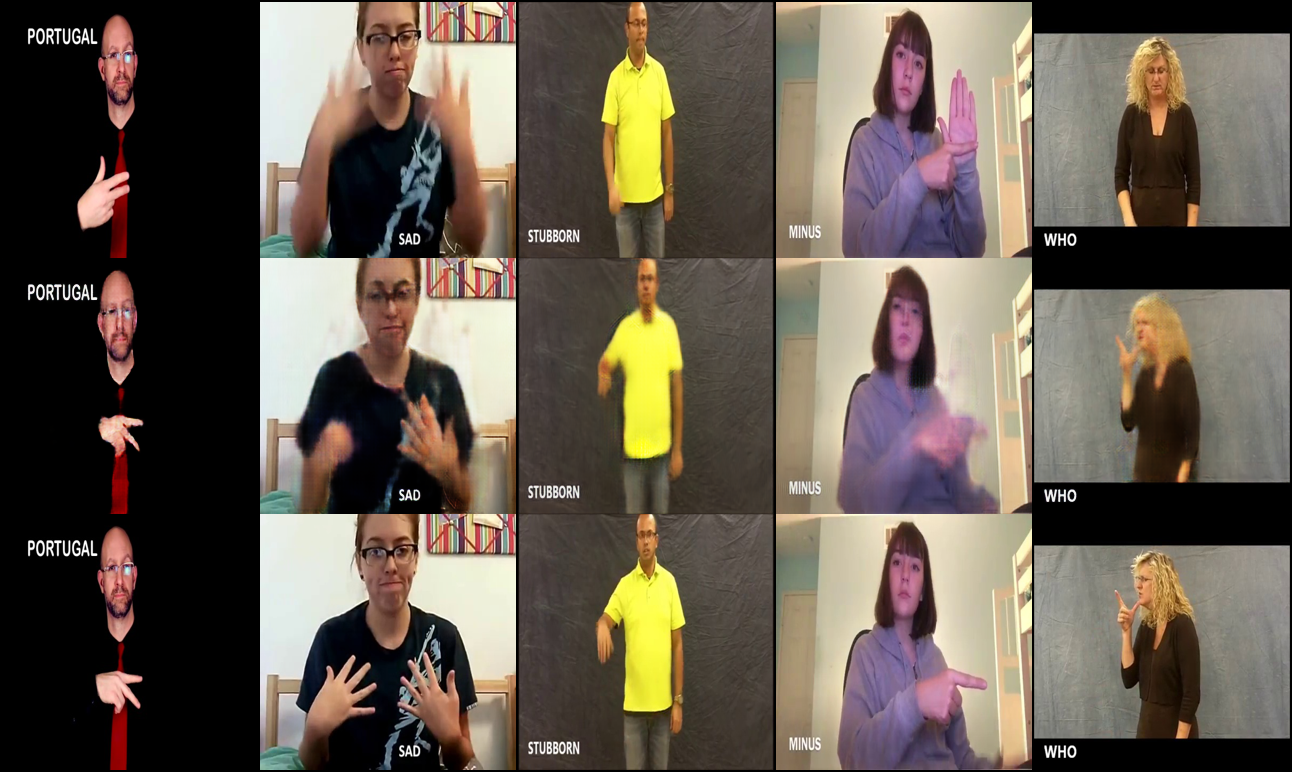}
  \caption{Five examples of our produced sign language glosses. First line shows the input image, second the image produced by our SignlanguageGAN, third the ground truth.}
  \label{fig:results_grid_1}
\end{figure}

Since SignLanguageGAN is the first SLP approach to use the MS-ASL dataset it is difficult to make a meaningful comparison. Nevertheless, we use a subset of the MS-ASL dataset with 184 videos consisting of a total of 16601 frames as a benchmark 
resulting in a PSNR of 26.044, MSE of 260.363 and SSIM of 0.893 after 300 epochs of training.
We compare our results to the results for three individual signers from Text2Sign~\cite[p. 901 Table 3]{Text2Sign}, even though our test setup is significantly different from the Text2Sign setup. First, we did not have access to the SMILE dataset, which makes the comparison much more difficult and also prevents us from using the pix2pix HD~\cite{Pix2PixHD} architecture. Second, Text2Sign performs a fine-tuning for the three signers.
Since pix2pix HD performs better and SLP for only three signers is not as difficult as for more than 200 we are not surprised that Text2Sign results, with an average PSNR of 25.604, MSE of 166,614 and SSIM of 0.942 are better than ours in terms of pure numbers. Nevertheless, we show that our architecture performs well on extremely diverse data that also contains many low quality samples and without special fine-tuning for individual signers. Figure~\ref{fig:results_grid_1} shows five examples of our results.

\section{Conclusion}

In this paper, we present the SignLanguageGAN, a novel approach for photo-realistic synthesis of sign language videos. 
In addition to previous work on SLP we extended our GAN with a human semantic parser.
This allows our system to learn region-level spatial layouts in order to guide the generator to produce more realistic appearing videos on a large variety of signers.
The introduction of a periodic lambda for weighting the discriminator is worth to investigate further since it allows to create images with less artifacts in long training sessions.

\bibliographystyle{acl_natbib}
\bibliography{bibliography}

\end{document}